\documentclass[conference]{IEEEtran}
\IEEEoverridecommandlockouts
\usepackage{cite}
\usepackage{amsmath,amssymb,amsfonts}
\usepackage{algorithmic}
\usepackage{graphicx}
\usepackage{textcomp}
\usepackage{xcolor}
\usepackage{xspace}
\usepackage[utf8]{inputenc} 
\PassOptionsToPackage{hyphens}{url}
\usepackage{hyperref} 
\usepackage[hyphenbreaks]{breakurl}

\hypersetup{
	breaklinks=true,   
	colorlinks=true,   
	pdfusetitle=true,  
}

\def\BibTeX{{\rm B\kern-.05em{\sc i\kern-.025em b}\kern-.08em
    T\kern-.1667em\lower.7ex\hbox{E}\kern-.125emX}}

\makeatletter
\DeclareRobustCommand\onedot{\futurelet\@let@token\@onedot}
\def\@onedot{\ifx\@let@token.\else.\null\fi\xspace}

\def\eg{\emph{e.g}\onedot}

\def\wrt{w.r.t\onedot}

\def\etal{\emph{et al}\onedot}

\begin{document}

\title{Compliance Challenges in Forensic Image Analysis Under the Artificial Intelligence Act\\
\thanks{Work was supported by Deutsche Forschungsgemeinschaft (DFG, German Research Foundation) as part of the Research and Training Group 2475 "Cybercrime and Forensic Computing" (grant \#393541319/GRK2475/1-2019).}
}

\author{\IEEEauthorblockN{Benedikt Lorch\IEEEauthorrefmark{1}, Nicole Scheler\IEEEauthorrefmark{2}, Christian Riess\IEEEauthorrefmark{1}}
\IEEEauthorblockA{\IEEEauthorrefmark{1}\textit{IT Security Infrastructures Lab}, \textit{Friedrich-Alexander-Universität}, Erlangen, Germany}
\IEEEauthorblockA{\IEEEauthorrefmark{2}\textit{International Criminal Law Research Unit}, \textit{Friedrich-Alexander-Universität}, Erlangen, Germany\\\{benedikt.lorch, nicole.scheler, christian.riess\}@fau.de}
}

\maketitle

\begin{abstract}
	In many applications of forensic image analysis, state-of-the-art results are nowadays achieved with machine learning methods. However, concerns about their reliability and opaqueness raise the question whether such methods can be used in criminal investigations. So far, this question of legal compliance has hardly been discussed, also because legal regulations for machine learning methods were not defined explicitly.
	
	To this end, the European Commission recently proposed the artificial intelligence (AI) act, a regulatory framework for the trustworthy use of AI. Under the draft AI act, high-risk AI systems for use in law enforcement are permitted but subject to compliance with mandatory requirements.
	
	In this paper, we review why the use of machine learning in forensic image analysis is classified as high-risk. We then summarize the mandatory requirements for high-risk AI systems and discuss these requirements in light of two forensic applications, license plate recognition and deep fake detection. The goal of this paper is to raise awareness of the upcoming legal requirements and to point out avenues for future research.
\end{abstract}


\section{Introduction}
	Forensic image analysis subsumes a range of techniques to analyze the contents of an image or the image itself in legal matters~\cite{SWGDE2017ForensicImageAnalysis}. Relevant sub-disciplines include photographic comparison, photogrammetry, source identification, and image authentication. The initial motivation for the development of these fields was to support criminal investigations by providing scientifically grounded statements about probative image material. Although forensic image analysis has now attracted the interest of a wider audience such as journalistic fact-checkers and insurance companies, the analysis of evidence in criminal investigations is still one of the field's core interests.
	
	After the recent success of deep learning, the research focus in multimedia forensics shifted from analytical models to machine learning. Machine learning is attractive for forensic image analysis for several reasons. First, advances in object detection, image denoising, and body pose estimation oftentimes directly translate to forensic tasks like license plate recognition~\cite{Moussa2021Sequence}, noise residual extraction~\cite{Cozzolino2020Noiseprint}, and photogrammetric height estimation~\cite{Thakkar2021ForensicHeightEstimation}. Second, many forensic traces are notoriously difficult to model analytically, due to limited knowledge about the image origin (\eg, camera hardware or processing) and due to the complex interaction between traces and image transformations. In these cases, learning from large sets of examples provides an appealing solution. Additionally, machine learning methods excel in broader application scenarios, whereas the traditional analytical methods are usually confined to a narrow application scope~\cite{Kraetzer2015Benchmarking}.
	
	Despite their high performance, machine learning methods suffer from a number of intrinsic shortcomings. Most importantly, their opaqueness makes it difficult to explain how the method works internally and to trace a particular output. Hence, an open challenge is how to audit and verify that a machine learning system works as intended and has no adverse side effects or biases.
	Additionally, machine learning methods are sensitive to training-test mismatches. This issue is particularly relevant in forensic applications, where methods are regularly exposed to images from uncontrolled origins, which may not be covered by the training data. In such cases, machine learning methods are prone to making wrong predictions~\cite{Lorch2020BayesianLogisticRegression}. Especially in applications where methods rely on imperceptible traces, detecting such mismatches is difficult.
	
	These limitations raise the question to what extent machine learning tools for forensic image analysis can be applied to criminal prosecution cases, in all stages from the investigation phase to providing expert witness testimony in the courtroom. Although the use of clues from forensic image analysis in the investigation phase may appear less delicate, such clues can lead to further investigative measures that affect fundamental rights of the suspect. Investigative measures that restrict the fundamental rights of an individual must be based on a sufficient and secured factual basis.
	Therefore, forensic analysis methods must be as reliable as possible, and their legal compliance cannot be taken for granted.
	
	Although current research focuses on learning-based methods, their practical use in law enforcement and their legal compliance are rarely discussed. This may be because legal compliance regulations either vary between jurisdictions, are unclear to the technical community, or even do not exist yet.
	
	In April 2021, the European Commission proposed a regulatory framework for the trustworthy use of AI, the \emph{artificial intelligence act}~(AIA)~\cite{EuropeanCommission2021ArtificialIntelligenceAct}.\footnote{The AIA's definition of AI technologies includes machine learning and deep learning, but also knowledge-based reasoning, expert systems, statistical approaches, and search and optimization methods. The Commission deliberately chose this broad definition of AI to include future developments~\cite{Spindler2021KIVerordnung}.} Although the proposal is still under discussion, the AIA will ultimately regulate the use of AI methods across the EU, including in law enforcement.
	Therefore, the AIA establishes legal clarity on which uses of AI are prohibited, which uses are permitted subject to mandatory safeguards, and which uses are permitted with few or no restrictions.
	
	In this paper, we examine the AIA and analyze its impact on learning-based forensic image analysis methods. The AIA classifies forensic image analysis applications in law enforcement as high-risk. Under the AIA, AI systems for these applications are permitted but they must meet mandatory requirements.
	As our contribution, we summarize and discuss the most relevant requirements in light of two forensic applications, namely license plate recognition and deep fake detection. The goal of this paper is to raise awareness within the research community about legal requirements for AI tools, and to point out directions for future research motivated by the AIA.
	
	For our discussion we choose these two applications, because machine learning is a promising approach in both cases, far outperforming traditional methods and humans. While license plate recognition has immediate application in law enforcement, deep fake detection is still a young field, but nevertheless highlights some of the key challenges towards making learning-based tools legally compliant under the AIA.
	
	This paper is organized as follows: Section~\ref{sec:related_work} reviews related work on legal requirements in forensic image analysis. Section~\ref{sec:artificial_intelligence_act} summarizes the AIA and the requirements to high-risk AI systems. In Sec.~\ref{sec:conformity_challenges}, we discuss these requirements using the two forensic applications of license plate recognition and deep fake detection. Section~\ref{sec:conclusion} concludes the paper.
	

\section{Related Work}
	\label{sec:related_work}
	Only few works discuss the applicability of forensic algorithms from a legal perspective. As a notable example, Kraetzer and Dittmann summarize the U.S.\ Federal Rules of Evidence and the Daubert criteria and discuss the maturity of camera device forensics (PRNU) in light of these compliance rules~\cite{Kraetzer2015Benchmarking}.
	In a similar spirit, De Rosa \etal summarize ISO standards and guidelines related to digital investigations and discuss application scenarios of traditional media forensics methods~\cite{DeRosa2014Standardization}. However, neither these ISO standards nor the Daubert criteria give concrete advice about AI methods. Moreover, the Daubert criteria only apply to expert witness testimony in the U.S., but not during the investigation phase.
	
	Furthermore, the Scientific Working Group on Digital Evidence (SWGDE) and the European Network of Forensic Science Institutes (ENFSI) released a series of guides to promote best practices and to establish comparability in forensic casework. These documents cover, for example, image authentication, forensic image comparison, and requirements for testing forensic tools. However, these documents do not yet consider the use of AI, with one exception: In~\cite{SWGDE2021AIVideoAnalysis}, the SWGDE outlines the emerging use cases in law enforcement and the challenges related to AI-based video analysis. While acknowledging the potential benefits of AI tools, the guide cautions investigators to be aware of the limitations of AI and to diligently follow established standards.
	Similar to best practice guides, forensic process models guide investigations and facilitate comparability and reproducibility~\cite{Siegel2021Handcrafted}. However, both these guides and process models do not yet take into account the particular challenges related to AI methods.
	
	In the related field of biometrics, the use of AI for facial recognition has raised several concerns, including the risk of discrimination, privacy aspects of large facial databases, and the potential use for public surveillance~\cite{Rueckert2021FacialRecognition}. Because these concerns outweigh the benefits, the AIA prohibits facial recognition systems for public surveillance, with special exceptions for law enforcement. In this paper, however, we focus on other applications of forensic image analysis such as license plate recognition and deep fake detection. For a detailed discussion of facial recognition, we refer to the in-depth analysis by the European Parliament Research Service~\cite{Madiega2021FacialRecognition}.


\section{Artificial Intelligence Act}
	\label{sec:artificial_intelligence_act}
	This section first provides general background on the AIA. Then we discuss why forensic image analysis applications used by law enforcement are classified as high-risk. Lastly, we summarize the requirements to high-risk AI systems.

	\subsection{Background}
		In the AIA, the European Commission proposes a horizontal legal framework for the trustworthy use of AI~\cite{EuropeanCommission2021ArtificialIntelligenceAct}.
		This proposal is based on recommendations by an appointed high-level expert group~\cite{HighLevelExpertsGroup2019Trustworthy}, a preceding white paper, and public stakeholder consultation. The AIA applies to both private, public, as well as extraterritorial providers whose AI tools are used in the EU.
	Once adopted (presumably in 2024 at the earliest~\cite{Veale2021Demystifying}), the AIA will become directly applicable law in all EU member states.
		
		The AIA classifies the use of AI in four risk categories called unacceptable risks, high risks, transparency risks, and minimum risks. Applications with unacceptable risk are prohibited. High-risk AI systems are permitted but subject to mandatory requirements. Applications with transparency risks or minimum risks are permitted subject to transparency obligations or without restrictions, respectively. This paper focuses on the high-risk category, in which we see many forensic image analysis techniques when used by law enforcement.
		
		The AIA leaves the technical implementation of these requirements to the providers of the respective AI systems, so that providers can take into account the scientific and technical progress. Nevertheless, the Commission will instruct the European standardization organizations to translate these requirements into harmonized technical standards. If providers adhere to these standards, their AI systems are given a presumption of conformity (Art.~40). These harmonized standards will further clarify and technically define some of the still abstract requirements stated in the AIA. Nevertheless, we believe that the research community benefits from an early discussion of these requirements, such that these requirements can be taken into consideration in future research.

	
	\subsection{High-risk AI systems}
		Annex III explicitly states law enforcement application scenarios that the Commission classifies as high-risk. These include AI systems for evaluating the reliability of evidence in the course of investigation or prosecution of criminal offenses, and AI systems for deep fake detection.
		Therefore, we conclude that AI-based image authentication counts as high-risk application.

		It is not explicitly stated whether AI systems for source identification or photographic comparison, including license plate recognition, are considered high-risk. However, the Commission is entitled to expand the list of high-risk applications. AI systems can be added to the list when they pose a similar or even higher risk to the health, safety, or fundamental rights than the applications already listed. Since source identification and photographic comparison aim at identifying a suspect, potentially leading to investigative measures against the suspect, we expect these applications to also be considered high-risk.
		
	\subsection{Requirements to high-risk AI systems}
		\label{ssec:requirements}
		We now summarize requirements to high-risk AI systems. We focus on requirements that are relevant to the development of the AI model, leaving aside operational requirements.
		
		\subsubsection{Accuracy, robustness and cybersecurity (Art.~15)}
			High-risk AI systems must achieve an appropriate level of accuracy, robustness, and cybersecurity for their intended purpose. AI systems must tolerate errors, faults and inconsistencies that may appear within their working environment. AI systems must be resilient to attacks, in particular related to manipulations of the training dataset (data poisoning), misleading inputs (adversarial examples), and model flaws.
		
		\subsubsection{Risk management and testing (Art.~9)}
			The known and foreseeable risks associated to the AI system should be identified, evaluated, and documented systematically within a risk management system. To alleviate these risks, mitigation strategies should be implemented. High-risk AI systems must be tested to ensure they perform consistently for their intended purpose and to validate the risk mitigation strategies.
			
		\subsubsection{Data and data governance (Art.~10)}
			Training, validation, and testing data are subject to appropriate data governance practices. These practices concern the data collection and pre-processing, data assumptions, data availability and quantity, and the examination of possible biases, among other things. In particular, all data sets must be relevant, representative, free of error, and complete. Datasets should take into account the characteristics of the individuals and the geographical, behavioral, or functional settings, for which the AI system will be used.
			
		\subsubsection{Transparency (Art.~13) and human oversight (Art.~14)}
			The AI system must be sufficiently transparent such that users can interpret and use the AI system's output appropriately. The degree of transparency must enable the user to oversee the AI system in order to prevent and minimize risks to health, safety, and fundamental rights. Providers must specify measures that allow a human operator to fully understand the capabilities and limitations of the AI system, to correctly interpret the AI system's output with the tools available, to decide to disregard the AI system's output, and to remain aware of human tendency to over-rely on the AI system.
			
		\subsubsection{Documentation}
			The AI system must be accompanied by several sources of documentation including an instruction manual (Art.~13), a technical documentation (Annex~IV), a risk management system (Art.~9), and a quality management system (Art.~17). The instruction manual specifies the AI system's capabilities and limitations. This includes its intended purpose, the expected level of accuracy, robustness and cybersecurity, known and foreseeable circumstances that impact these performance metrics, specifications for the input data, and the interpretability measures. The technical documentation describes in detail the system components and its development including conceptional decisions. The risk management system describes the risks associated with the use of the system and how they are mitigated. The quality management system documents compliance with all regulations of the AIA.
			
\section{Compliance Challenges}
	\label{sec:conformity_challenges}
	In this section, we review the requirements from Sec.~\ref{ssec:requirements} and discuss to what extent these requirements are currently addressed in research. We specifically use the examples of license plate recognition and deep fake detection.
	
	\subsubsection{Accuracy, robustness, cybersecurity}
		In simple terms, this requirement demands AI systems to perform well. Overall, we find that these requirements align with the goals and established performance metrics in the research community.
		
		For example, recent work on license plate recognition aimed at increasing accuracy and improving robustness to image degradations compared to prior work~\cite{Moussa2021Sequence}. We are not aware of studies on the adversarial robustness of forensic license plate recognition. However, the input images usually stem from security cameras with restricted access. Therefore, a feasible direction to prevent attacks could be to actively embed security schemes for image authentication. Nevertheless, a more common issue than tampering are image degradations due to environmental aspects such as rain, dirt, and difficult lighting conditions. Future work could study the robustness \wrt these degradations.
		
		In deep fake detection, a recent line of work focuses on improving the accuracy and robustness in light of unseen manipulations. Several works also studied the resilience of deep fake detectors to evasion attacks~\cite{Verdoliva2020Overview}.
		Despite these developments, improving robustness to real-world image degradations and unknown manipulations, and defending against attacks are still open problems, but are being actively researched.
		
	\subsubsection{Risk management and testing}
		The AIA requires providers to thoroughly test their AI systems. Because AI systems do not give theoretical guarantees, testing is essential to establish the required performance metrics, to identify risks, and to validate risk management strategies. So far, there are no established or standardized protocols for AI testing~\cite{ExamAI2021}.
		
		Although technical research papers usually include an experimental validation, these validations are often limited in breadth and depth, they only focus on a particular performance metric, and testing protocols vary between papers. In contrast, testing AI systems for real-world applications requires a structured testing protocol. This testing protocol should assert, in addition to performance, quality criteria such as fairness, safety and reliability. Yet, this testing protocol must be sufficiently flexible to support different notions of these quality criteria, because the notions of fairness and safety vary between different applications.
		
		To this end, the \emph{ExamAI} project~\cite{ExamAI2021} proposed \emph{assurance cases} as an argumentation framework to structurally document how an AI system meets the required quality characteristics. In a nutshell, assurance cases consist of a normative claim, arguments to support the claim, and factual evidence to support these arguments. Assurance cases therefore make transparent to both developers and external auditors what has been experimentally validated and how this factual evidence supports the claim of these qualities. As long as there are no standardized protocols, we see assurance cases as a helpful tool.
		
		However, we note that large-scale testing is likely beyond the capacity of individual researchers and requires assistance from companies, research institutes, or federal agencies. Nevertheless, the required risk management system provides motivation to research structured testing protocols (\eg assurance cases for license plate recognition and deep fake detection), to identify risks associated to forensic AI applications, and to develop mitigation strategies. For example, future work on detecting deep fakes could investigate the risk of discrimination, where first results showed that detectors tend to misclassify people with skin diseases and overly smooth faces~\cite{Prabhu2020Discrimination}.
		
	\subsubsection{Data and data governance}
The AIA prescribes training, validation, and testing datasets to be relevant, representative, free of error, complete, and datasets must be examined for biases. Arguably, the meaning of these properties must be discussed in the context of each application.
		
		In the context of license plate recognition, relevance could mean that all images must show license plates, and error-free means that labels are correct. Representativity and completeness can be interpreted such that images must cover all types of image degradations and geographical variations relevant for the actual use case. A typical source of bias could be an over- or under-representation of some characters or a geographic region.
		Unfortunately, publicly available license plate datasets are small, which limits their representativeness and makes them too incomplete to reflect real-world conditions. For this reason, recent works trained their AI models with synthetic images~\cite{Moussa2021Sequence, Rossi2021UnetLPR}. While synthetic training data mitigates some of these issues, synthetic images must be representative enough such that the model generalizes to real images.
		
		The current trend towards larger models creates demand for even larger datasets. Because data collection is very expensive, compiling a complete and representative dataset is, again, likely beyond the capacity of individual researchers. Although scraping images from the internet is an appealing shortcut, such datasets are likely to contain biases or labeling errors.
		
		We welcome large datasets provided by governmental organizations, \eg in the DARPA MediFor program, or private organizations, \eg Facebook's deepfake detection challenge dataset. This latter dataset features a wide variety of gender, skin, tone, ethnicity, age, and other characteristics, and it contains a large number of image degradations. Nevertheless, these datasets can still contain biases and their large size makes them difficult to screen.
		
		An interesting direction for future research could therefore be to develop automatic screening tools and metrics to assess the quality of datasets. As an example, recent work identified numerous labeling errors in several popular datasets including ImageNet through an automated method~\cite{Northcutt2021Confident}. Another open problem is how to document that datasets meet the required properties. Future work could develop assurance cases to assert these data quality properties in a structured manner.
		
	\subsubsection{Transparency and human oversight}
		Interestingly, the AIA in its current draft requires only sufficient transparency, while the high-level expert group called for a stronger notion of explainability, namely that humans can understand and trace predictions by an AI system~\cite{HighLevelExpertsGroup2019Trustworthy}. This weakened requirement leaves opportunities for the use of black-box models such as neural networks.
		Nevertheless, the degree of transparency achieved by the AI system must enable humans to correctly interpret and potentially disregard the AI system's output. To our understanding, this ability to disregard the AI system's output requires a way to identify malfunctioning. To this end, we see two complementary strategies: explaining model predictions and uncertainty-aware detectors.
		
		Work on \emph{explainable AI} (XAI) can be categorized into inherently interpretable models and post-hoc explanations. Two inherently interpretable approaches for deep fake detection rely on hand-crafted features and use a random forest~\cite{Siegel2021Handcrafted} or kNN classifier~\cite{Matern2019VisualArtifacts}. These methods perform well when the forensically relevant traces are strong, as it is the case in constrained scenarios. But when forensically relevant traces are weak, \eg statistical traces after resizing and compression, deep learning outperforms hand-crafted features~\cite{Verdoliva2020Overview}.
		For deep learning models, a few post-hoc explanation methods have been applied to deep fake detectors~\cite{Malolan2020DeepFakesXAI}, but so far only studied in very limited settings. In license plate recognition, first works studied saliceny maps~\cite{Lorch2019LicensePlates} and a sequential architecture with intermediate inspection possibilities~\cite{Rossi2021UnetLPR}.
		
		However, we share the opinion that current post-hoc XAI approaches are incapable of providing reliable explanations for every individual prediction~\cite{Rudin2019Interpretable, Ghassemi2021FalseHope}. This is because simple approximations may not be faithful to the original black-box model and because human interpretations based on post-hoc explanations may diverge from the actual model computation.
		Even though post-hoc XAI cannot provide explanations for every prediction, these visualizations can nevertheless indicate failure cases. For example, a deep fake detection result can be deemed unreliable if a saliency map points at the background region even though the detection should focus on the foreground. Therefore, post-hoc XAI approaches can at least unveil some cases of malfunctioning.
		Given that the AIA does not specify the required level of interpretability, it is worth exploring both directions: stronger inherently interpretable models and post-hoc explanations. We note that explaining AI methods for forensic tasks may be more difficult than for image recognition, because the forensically relevant traces are often imperceptible. Therefore, an open challenge is to visualize the meaning of these traces even to experts.

		A complementary approach to identifying erroneous predictions could be uncertainty-aware methods, \eg through Bayesian modeling~\cite{Lorch2020BayesianLogisticRegression}. These methods provide an uncertainty with every prediction. This enables the operator to quantify how much to trust in the prediction. So far, we are not aware of uncertainty-aware detectors for license plate recognition or deep fake detection. Despite first results, the reliability of these detectors has not been studied on a large scale.
		
		
	\subsubsection{Documentation}
		In research, technical papers sometimes fall short of discussing limitations of a proposed method, assumptions on the input data, and potential risks associated to an AI application. We see this documentation requirement as a motivation to increase transparency also in research papers.
		
		In fact, major conferences such as \emph{NeurIPS 2021} introduced a paper checklist for responsible machine learning research.\footnote{\url{https://neuripsconf.medium.com/introducing-the-neurips-2021-paper-checklist-3220d6df500b}, accessed on 16.02.2022} Interestingly, this checklist overlaps with the AIA documentation requirements in terms of discussing limitations and assumptions, identifying potential risks with negative societal impact, and requiring necessary technical details for reproducibility. At the same time, reviewers are instructed to acknowledge authors for discussing limitations and societal implications.
		Similarily, IEEE T-IFS published guidelines for deep learning submissions\footnote{\url{https://signalprocessingsociety.org/sites/default/files/uploads/publications_resources/docs/Guidelines__deep_learning_submissions.pdf}, accessed on 16.02.2022}, although these only focus on reproducibility, leaving aside data governance or risks.
		
		After all, improving transparency in research papers would not only foster reproducibility. It would also allow the research community to understand potential risks, to develop mitigation strategies, to establish best practices in data governance, and to understand limitations of state-of-the-art methods.
		
\section{Conclusion}
	\label{sec:conclusion}
	Recently proposed methods in forensic image analysis are increasingly based on machine learning. Despite being motivated by law enforcement applications, the practical use of AI tools and their legal compliance are rarely discussed. To this end, the AIA establishes legal clarity that AI systems in such high-risk applications are permitted but subject to mandatory requirements. In this paper, we reviewed these requirements and discussed their alignment with currently acknowledged goals of the research community.
	Although it is the providers of AI systems who need to show compliance with the AIA's regulations, some of these requirements present interesting directions for future research, \eg identifying risks and mitigation strategies, developing data governance practices, and improving interpretability.
	While the AIA is still subject to amendment, we hope this paper will stimulate discussion about the practical applicability of AI tools in multimedia forensics.
	
\bibliographystyle{IEEEtran}
\bibliography{artificial_intelligence_act}

\end{document}